%% file: main.tex
\documentclass{article}
\usepackage{spconf,amsmath,graphicx}
\usepackage{color}
\usepackage{multirow}
\usepackage{upgreek}
\usepackage{amsfonts} 
\usepackage{bbm}
\usepackage[utf8]{inputenc}
\usepackage{enumitem}
\setlist{nosep, leftmargin=14pt}

\usepackage{mwe}

\definecolor{cb_orange}{rgb}{1.0,0.51,0.0}
\definecolor{cb_blue}{rgb}{0.22,0.49,0.72}
\definecolor{cb_green}{rgb}{0.3,0.67,0.29}
\definecolor{cb_red}{rgb}{0.89,0.1,0.11}
\definecolor{cb_pink}{rgb}{1, 0, 0.4}

\title{Scribble-Supervised Cell Segmentation using \\Multiscale contrastive Regularization}
\name{Hyun-Jic Oh $^{1}$ \qquad
 Kanggeun Lee $^{2}$ \qquad
 Won-Ki Jeong $^{1,*}$ \qquad \thanks{*Corresponding author: wkjeong@korea.ac.kr}
}
\address{
$^{1}$ Department of Computer Science and Engineering, Korea University,  Seoul, Korea \\
$^{2}$ School of Electrical and Computer Engineering, UNIST,  Ulsan, Korea
}

\begin{document}
\maketitle{}

\begin{abstract}
Current state-of-the-art supervised deep learning-based segmentation approaches have demonstrated superior performance in medical image segmentation tasks.
However, such supervised approaches require fully annotated pixel-level ground-truth labels, which are labor-intensive and time-consuming to acquire.
Recently,~\textit{Scribble2Label} (S2L) demonstrated that using only a handful of scribbles with self-supervised learning can generate accurate segmentation results without full annotation. 
However, owing to the relatively small size of scribbles, the model is prone to overfit and the results may be biased to the selection of scribbles.
In this work, we address this issue by employing a novel multiscale contrastive regularization term for S2L. 
The main idea is to extract features from intermediate layers of the neural network for contrastive loss so that structures at various scales can be effectively separated.
To verify the efficacy of our method, we conducted ablation studies on well-known datasets, such as Data Science Bowl 2018 and MoNuSeg. 
The results show that the proposed multiscale contrastive loss is effective in improving the performance of S2L, which is comparable to that of the supervised learning segmentation method.

\end{abstract}

\begin{keywords}
Contrastive Learning, Weakly Supervised Learning, Cell Segmentation
\end{keywords}

\input{Introduction}

\input{Method}

\input{Experiment}

\input{Conclusion}
\section{Compliance With Ethical Standards}

This research study was conducted retrospectively using human subject data made available in open access by Data Science Bowl~\cite{Caicedo2019} and MoNuSeg~\cite{monuseg}. Ethical approval was not required as confirmed by the license attached with the open access data.

\section{Acknowledgments}
This work is supported by the National Research Foundation of Korea (NRF-2019M3E5D2A01063819, NRF-2021R1A6\\A1A13044830), the Institute for Information \& Communications Technology Planning \& Evaluation (IITP-2022-2020-0-01819), and the Korea Health Industry Development Institute (HI18C0316).

\bibliographystyle{IEEEbib}
\bibliography{refs}

\end{document}

%% file: Introduction.tex
\section{Introduction}
\label{sec:intro}
Cell segmentation~\cite{Cell_seg} is a fundamental and important image processing method that is necessary in many biomedical image analysis tasks, such as cell counting and cell morphology analysis.
Supervised learning with convolutional neural networks (CNNs) has shown superior performance in cell segmentation, but it requires a sufficient number of manually annotated pixel-level labels for robust results, which demands a labor-intensive and time-consuming label generation process. 
In addition, even the labels generated by experts are not perfect and may contain errors, especially in complex or subtle boundaries, which impairs the performance of supervised learning-based approaches. 

To alleviate the need for manual annotations, 
various weakly supervised segmentation algorithms that do not use full ground-truth labels have been introduced. 
Bearman \textit{et al.}~\cite{Bearman2016whatsthepoint} proposed a weakly supervised segmentation scheme that learns from point annotations and generic objectness prior~\cite{alexe2012measuring}, providing the guidelines for separating foreground and background. 
More recently, learning from point annotations~\cite{yoo2019pseudoedgenet, Chen2021, Zhao2021} in the medical image domain showed the potential of weakly supervised learning, which closes the performance gap to fully supervised learning.
Another approach is to use sparse scribbles rather than per-object point annotation.
Lin \textit{et al.}~\cite{Lin_2016_CVPR} used
a super-pixel-based pseudo-label generated by scribble annotations.
Scribble2Label (S2L)~\cite{S2L} proposed scribble-based supervised learning with pseudo-labeling on an unlabeled area. 
Even though S2L achieved robust results, it still has a limitation of overfitting scribble annotations, which may impair the segmentation performance.

The main motivation of this work is to address the limitation of S2L by introducing a new regularization energy, which stems from the observation of recent advances in contrastive learning.  
Recently, self-supervised contrastive learning (SimCLR)~\cite{simCLR} has shown notable performance as a significant variant of self-supervised learning. 
The main intuition behind SimCLR is that transformation of similar images should have similar representations 
(i.e., latent features of a neural network), while different images should have different representations in the feature space.
Khosla \textit{et al.}~\cite{Khosla2020} demonstrated the possibility of the scalability of self-supervised contrastive learning to supervised contrastive learning in an image classification task. 
Supervised contrastive learning aims to push the anchor far away from samples in different classes, and conversely, to pull the anchor close to samples in the same class. 
However, existing contrastive learning works mostly focus on image classification. 
Contrastive learning in image segmentation is in its early stage, and only a few works have been attempted on image segmentation ~\cite{Zhong_2021_ICCV, chaitanya2020contrastive}. 

\input{overview}

In this study, we propose multiscale contrastive regularization based on pseudo supervision produced by scribble-supervised learning to improve the performance of cell segmentation.
The main contributions of our work are summarized as follows:
\begin{enumerate}
\item We introduce a novel adaptation of supervised contrastive learning from image classification into a weakly supervised pixel-level image segmentation. 
\item We introduce a generalized multiscale contrastive regularization applicable to weakly supervised learning by employing supervised contrastive regularization on various image resolutions. 
\item We demonstrate the efficacy of the proposed method by extending S2L and performing quantitative and qualitative analyses on well-known datasets (e.g., Data Science Bowl 2018 and MoNuSeg).

\end{enumerate}

%% file: overview.tex
\begin{figure*}[t]
\label{fig:overview}
\centering
\includegraphics[width=0.9\textwidth]{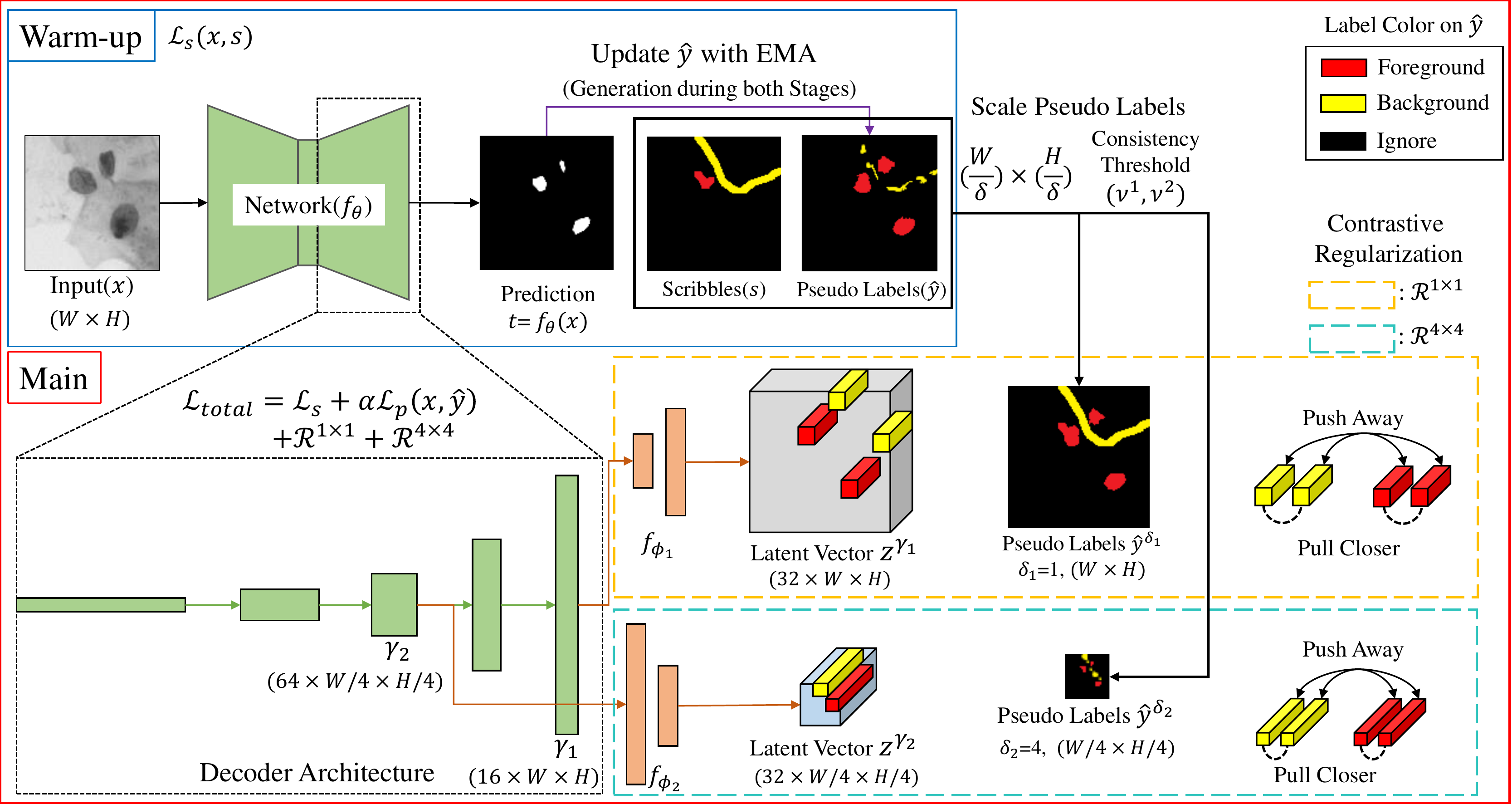}
\caption{
An overview of the proposed method. Because our method is based on S2L, we briefly described it~\cite{S2L} with the proposed method. 
First, the warm-up stage proceeds scribble-supervised learning and pseudo label generation. Next, while continuing the same process on main stage, we additionally conducts pseudo-label based learning and multiscale contrastive regularization. 
Putting each output from decoder layer $\gamma$ into projection head $f_\phi$, we obtain latent vector $z^{\gamma}$.
Here, we use scribbles and the generated pseudo labels to annotate the label of the latent vector.
According to the labeling standard $\nu$ and the resolution factor $\delta$, $\hat{y}^\delta$ is generated. With the assigned labels and latent vectors, we compute contrastive loss.
}
\end{figure*} 

%% file: Method.tex
\section{Method}

\subsection{S2L Overview}

The original S2L~\cite{S2L} consists of two stages: warm-up and main stages.
As described in Fig.~\ref{fig:overview}, in the warm-up stage, the CNN only learns from a set of scribbled pixels $\Omega$, and periodically generates pseudo labels through a filtering method based on the consistency threshold and exponential moving average (EMA) method. We rewrote the scribble-based supervised binary cross-entropy loss $\mathcal{L}_\text{s}$~\cite{S2L} as follow:
\begin{align}
    \mathcal{L}_\text{s}(x,s) = \frac{1}{|\Omega|}\sum_{i\in\Omega}{\text{BCE}(s_i,t_i)}
    \label{eq:supervised_loss}
\end{align}
where $x$ is an input image, $t=f_\theta(x)$ is a vectorized prediction generated by an arbitrary CNN $f_\theta$, and $s_i$ is a vectorized scribble label indexed by $i\in\Omega$.
Second, in the main stage, they~\cite{S2L} retrain the function $f_\theta$ with pseudo-labels~\cite{S2L} through the following main loss:
\begin{align}
    &\mathcal{L}_{\text{S2L}}(x,s,\hat{y}) = \mathcal{L}_{\text{s}}(x,s) + \alpha \mathcal{L}_{p}(x,\hat{y})
    \label{eq:main_loss}
    \\
    &\mathcal{L}_p(x,\hat{y}) = \frac{1}{|\Omega_p|}\sum_{i\in\Omega_p}{\text{BCE}(\hat{y}_i, t_i)}
    \label{eq:pseudo_loss}
\end{align}
where $\hat{y}$ is a vectorized pseudo-label and $\Omega_p$ is a set of pseudo-labels. We introduce the modified main stage in the next step as depicted in Fig.~\ref{fig:overview}.

\begin{figure*}[t]
\centering
\includegraphics[width=0.9\textwidth]{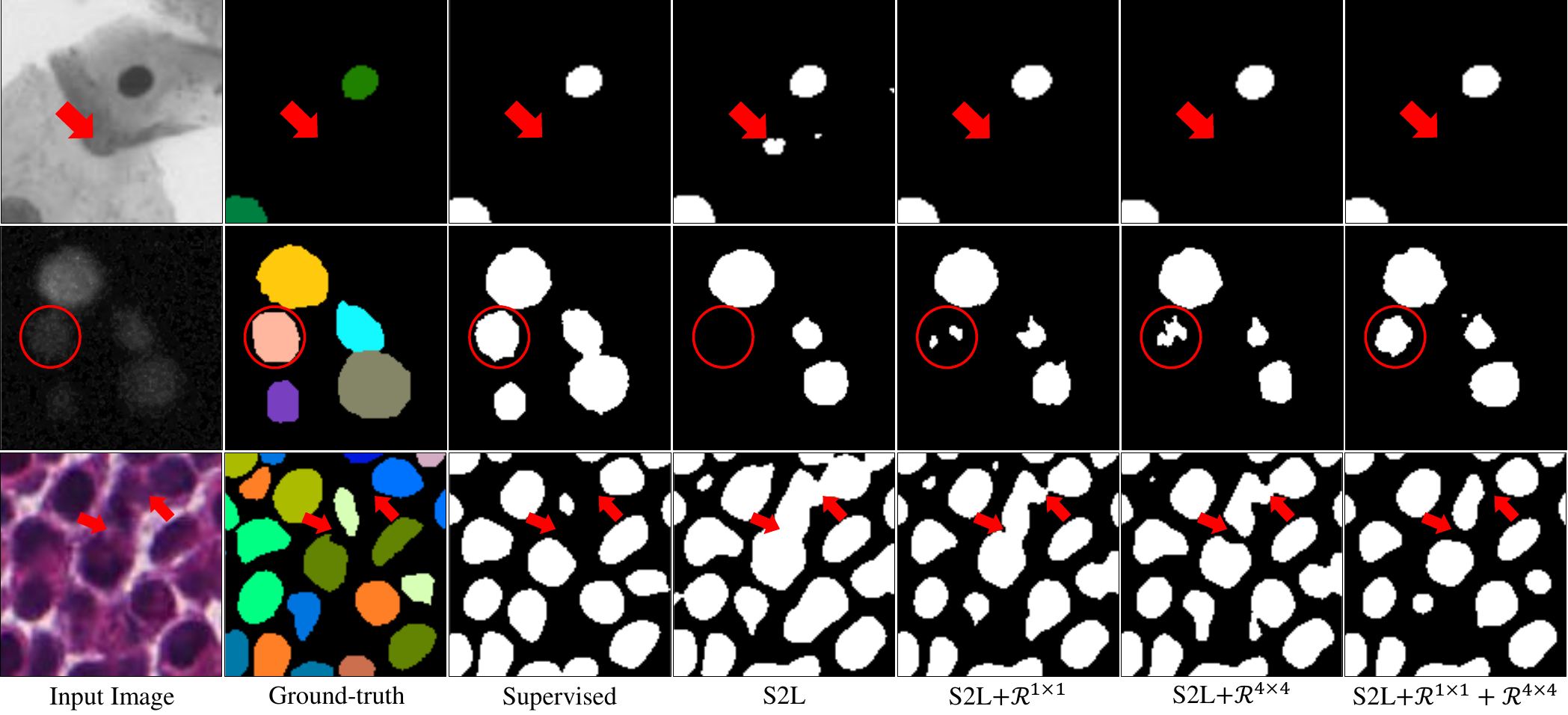}
\caption{Qualitative comparison between ground-truth, supervised, and our method (with various loss combinations).
Top row: DSB-BF, middle row: DSB-Fluo~\cite{Caicedo2019}, and bottom row: MoNuSeg~\cite{monuseg} datasets. For the input image of DSB-Fluo data, we increased the image contrast for better visualization.
Each ground truth image includes per-instance labels. 
Some notable regions are marked with arrows and circles.
}
\label{fig_qual}
\end{figure*}

\subsection{Multiscale Contrastive Regularization}
This section introduces multiscale contrastive regularization. 
We reformulate the main loss in Eq.~\ref{eq:main_loss} to employ multiscale contrastive regularization that restricts the convergence space of the arbitrary CNN $f$. 
Let $f_\theta(x)$ be composed of a sequence of transformations $f^{n-1} \circ ... \circ f^1 \circ f^0(x)$. We select a latent feature vector $v^\gamma = f^{\gamma}\circ...\circ f^{0}(x)$ such that $0 < \gamma \leq n$, and pass through a projection head similar to ~\cite{simCLR} as $f_\phi(v^\gamma) = z^\gamma$.
The projection head $f_\phi: \mathbb{R}^{C\times (HW/\delta^2)} \rightarrow \mathbb{R}^{C'^\times (HW/\delta^2)}$ 
consists of combinations of convolutional layers, batch normalization layers, and activation functions.
Now, we can define pseudo-supervised multi-scale contrastive regularization with respect to $\gamma$ and $\delta$ as:
\begin{align}
    \nonumber
    &\mathcal{L}_{c}(\hat{y},v;f_\phi,\gamma,\delta,\tau) = \\ &\dfrac{\delta^4}{W^2H^2}\sum_{i}\sum_{j}\text{BCE}(\hat{y}_i^\delta \wedge \hat{y}_j^\delta,\sigma{(\dfrac{z^\gamma_i}{||z^\gamma_i||_2} \cdot \dfrac{z^\gamma_j}{||z^\gamma_j||_2}/\tau))}
    \label{eq:MCR}
\end{align}
where $\sigma$ is a sigmoid function, $\cdot$ denotes the dot product, and $\wedge$ symbol denotes element-wise ``or" logical operator. And, 
\begin{equation}
    \hat{y}^\delta_i =
    \begin{cases}
      0, & \text{if}\ \dfrac{1}{\delta^2}||\hat{y}_{i:i+\delta^2}||_1 \leq \nu_1 \text{ or } i \in \Omega^{-} \\
      1, & \text{if}\ \dfrac{1}{\delta^2}||\hat{y}_{i:i+\delta^2}||_1 \geq \nu_2 \text{ or } i \in \Omega^{+}\\
      \text{Ignore}, & \text{otherwise} \\
    \end{cases}
    \label{eq:average_pool}
\end{equation}
In other words, $\hat{y}^\delta$ is a down-scaled vector of a pseudo-label generated by the factor $\delta \in \{2^i|i\in \mathbb{N}_0\}$ with consistency threshold values $\nu_1$ and $\nu_2$ for contrastive regularization at various resolutions. $i \in \Omega^{-}$ and $i \in \Omega^{+}$ denote the background and foreground scribble labels, respectively. In supervised contrastive learning, the softmax function can control a single positive sample with an anchor; however, it is an inefficient regularization in the segmentation task because of the large number of positive pixels.
Unlike the supervised contrastive regularization study~\cite{Khosla2020}, we employ a logistic function to scale the element-wise probability of pseudo-labels instead of a softmax function. The logistic function makes it feasible to handle multiple positive pixels for each sample $x$ simultaneously. 
Now, we can utilize supervised contrastive regularization in the latent space of various resolution features (i.e., it is extracted features on intermediate layers of encoder-decoder-like shaped model). 
Consequently, we can define the main loss function as a combination of the main loss of S2L with multiscale contrastive regularization as follows:
\begin{align}
    \nonumber
    \mathcal{L}_{total}(x,s,\hat{y},v;f_\phi,&\gamma,\delta,\tau) =
    \mathcal{L}_{\text{S2L}}(x,s,\hat{y}) \\ 
    \nonumber
    &+ \lambda_1\mathcal{L}_c(\hat{y}, v;f_{\phi_1},\gamma_1,\delta_1,\tau_1) \\
    &+ \lambda_2\mathcal{L}_c(\hat{y}, v;f_{\phi_2},\gamma_2,\delta_2,\tau_2) 
    \label{eq:total}
\end{align}
where $\alpha,\lambda_1,$ and $\lambda_2$ are weight parameters. 
Although Eq.~\ref{eq:MCR} can be extended to multiple regularization terms for all possible $\delta$, we empirically chose two regularization terms with different scale factors $\delta_1=1$ and $\delta_2=4$ in our experiment to balance between the training efficiency and the performance. 

%% file: Experiment.tex
\section{Experiment}
\input{result_table}
\subsection{Data}
We used three different cell image datasets. 
The first two sets of nuclei 2D images were obtained from BBBC038v1~\cite{Caicedo2019} known as Data Science Bowl (DSB) 2018 data.
In the DSB dataset, we used the Stage 1 training dataset with fully annotated masks, which were 16 bright-field (DSB-BF) 1000$\times$1000 images and 542 fluorescence (DSB-Fluo) images of various sizes.
We conducted 4-fold cross validation on these two datasets, so we splits
each dataset into 60\% for training set, 20\% for the validation set, and 20\% for the test set. 
The third dataset was MoNuSeg~\cite{monuseg}, which consisted of 30 training set
and 14 test set 1000$\times$1000 histopathology images. 
We conducted 5-fold cross validation and made predictions (inference) on the test set images. 
To manually draw scribbles, we referenced fully annotated masks. 

\subsection{Implementation Details}
Our baseline network is U-Net~\cite{unet} with five layers of decoder blocks, using the ResNet-34~\cite{He_2016_CVPR} backbone for the experiment. 
We used constant values of $\alpha=0.5$, $\lambda_1=0.5$, and $\lambda_2=10$.
For multiscale contrastive regularization, we empirically chose the third and last layers of the decoder (i.e., $\gamma_1$ and $\gamma_2$).
Let us denote $\lambda_1\mathcal{L}_c$ as $\mathcal{R}^{1\times1}$ and $\lambda_2\mathcal{L}_c$ as $\mathcal{R}^{4\times4}$. 
Because the loss operates more tolerant to similar latent vectors for the larger temperature parameter $\tau$ on contrastive loss~\cite{Wang_2021_CVPR},
we set the temperature $\tau_{1}=0.3$ and $\tau_{2}=0.1$. 
For the latent vector label allocation, we set $\nu_1=0$ and $\nu_2=1$ to use more reliable pseudo labels.
To reduce the cost of contrastive loss computation, we randomly sampled from each scale per class, setting 6000 as the maximum pixel number per class in pixel sampling. 
We used the same hyper-parameters used in the original S2L for Eq.~\ref{eq:main_loss}.

\subsection{Results}
For performance assessment, we compared our results with those of S2L and a supervised learning method using full annotation. 
Additionally, we conducted an ablation study to describe the segmentation performance of each regularization for a single resolution. 

For quantitative comparisons, 
we used the intersection of union (IoU) to evaluate the overall (semantic) segmentation accuracy and mean of the Dice-coefficient (mDice)~\cite{Nishimura2019} to measure the per-instance segmentation accuracy (see Table~\ref{tab:tab1}).
In DSB-BF and DSB-Fluo cases, multiscale contrastive regularization outperformed S2L and single-scale contrastive regularization. 
In the case of DSB-BF, single-scale contrastive regularization with $\mathcal{R}^{4\times4}$ showed the second highest performance. 
Because $\mathcal{R}^{4\times4}$ has a larger field-of-view, it has the advantage of separating regions that have similar intensity but structurally different. 
In DSB-Fluo, $\mathcal{R}^{1\times1}$ outperformed $\mathcal{R}^{4\times4}$ and showed a similar performance as the combination of $\mathcal{R}^{1\times1}$ and $\mathcal{R}^{4\times4}$. 
$\mathcal{R}^{1\times1}$ is more effective in distinguishing subtle intensity differences by comparing pixels in the highest resolution.
For the MoNuSeg data, $\mathcal{R}^{4\times4}$ performed best.

Fig.~\ref{fig_qual} shows the qualitative results with input, ground-truth, and the proposed method using various experimental settings. 
In the case of DSB-BF, S2L falsely predicted the nonnuclear region as the nucleus (see the arrows in the first row of Fig.~\ref{fig_qual}) that has a similar intensity as the nucleus. We believe this is owing to overfitting because scribbles are drawn inside cells, so the intensity is a dominant factor for the decision rather than the shape.
However, contrastive regularization distinguished the cell region correctly because different shapes could be separated by feature similarity across different scales.
In the case of DSB-Fluo, our proposed method detected the cell regions where S2L failed to detect. 
Note that our two scale-based regularization method (i.e., combining $\mathcal{R}^{1\times 1}$ and $\mathcal{R}^{4\times 4}$) detected the cell shape and boundary area that the other approaches failed to predict. 
In the case of MoNuSeg, S2L failed to separate cells nearby, which was improved by employing contrastive regularization. 

%% file: result_table.tex
\begin{table*}[t]
\centering
\renewcommand{\tabcolsep}{9pt}
\renewcommand{\arraystretch}{1.4}
\caption{Quantitative experiment results representing accuracy in the format of mDice / IoU. The first- and second-best results are highlighted in \textcolor{red}{red} and \textcolor{blue}{blue}, respectively.}
\label{table:1}
\begin{tabular}{|c|c|c|c|c|} 
\hline
 \textbf{Label} & \textbf{Method} & \textbf{DSB-BF} & \textbf{DSB-Fluo} & \textbf{MoNuSeg} \\ \hline\hline
 \multirow{4}{*}{Scribble} 
 & S2L\cite{S2L} 
 & 0.6980 / 0.8108 & 0.8235 / 0.8419 & 0.7098 / 0.6586
 \\ \cline{2-5}
 & S2L+$\mathcal{R}^{1\times1}$
 & 0.7068 / 0.8153 & \textcolor{blue}{0.8267 / 0.8565} & \textcolor{blue}{0.7165} / \textcolor{red}{0.6715}
 \\ \cline{2-5}
 & S2L+$\mathcal{R}^{4\times4}$  
 & \textcolor{blue}{0.7078 / 0.8292} & 0.8234 / 0.8505 & \textcolor{red}{0.7205} / 0.6603
 \\ \cline{2-5}
 & S2L+$\mathcal{R}^{1\times1}$+$\mathcal{R}^{4\times4}$
 & \textcolor{red}{0.7199 / 0.8368} & {\textcolor{red}{0.8268 / 0.8570}} & 0.7128 / \textcolor{blue}{0.6639}
 \\ \cline{2-5}
\hline\hline
 Full & Supervised & 0.7305 / 0.8665 & 0.8204 / 0.8790 & 0.7646 / 0.6867 \\ \hline
\end{tabular}
\label{tab:tab1}
\end{table*}

%% file: Conclusion.tex
\section{Conclusion}
In this study, we introduced supervised contrastive learning with multiscale features in scribble-based weakly supervised learning for cell segmentation. 
We demonstrated that two contrastive regularizers on different scales can achieve superior performance compared to S2L in DSB-BF and DSB-Fluo datasets, and a single-scale contrastive regularizer can also boost the results on the MoNuSeg dataset. 
We conclude that the proposed contrastive regularization is effective in avoiding the overfitting issue of scribble-based weakly supervised learning.

In future works, we plan to extend the current framework to adaptive contrastive regularization, which can leverage all the latent features from every resolution. 
Applying the proposed method to different image modalities, such as natural images, would be another interesting future research direction.